\documentclass[letterpaper, 10 pt, conference]{ieeeconf} 
\usepackage{xcolor}

\IEEEoverridecommandlockouts
\usepackage{cite}
\usepackage[font=footnotesize]{subfig}
\usepackage{multirow}
\usepackage{ifpdf}
\usepackage[pdftex]{graphicx}
\usepackage{epstopdf}
\usepackage[cmex10]{amsmath}
\usepackage{algorithmic}
\usepackage{array}
\usepackage{mdwmath}
\usepackage{mdwtab}
\usepackage{eqparbox}
\usepackage{verbatim}
\usepackage{xcolor}

\usepackage[font=footnotesize]{subfig}
\usepackage{fixltx2e}
\usepackage{stfloats}

\usepackage{amsmath,bm}
\usepackage{amssymb}
\usepackage{color}
\usepackage{subfig}

\usepackage{multirow}
\usepackage{float}
\usepackage{caption}
\usepackage{comment}

\usepackage{makecell}
\usepackage{amsmath}
\usepackage{graphicx}
\usepackage[colorlinks=true, allcolors=blue]{hyperref}
\usepackage{adjustbox}
\usepackage{blindtext}

\usepackage[font=footnotesize]{subfig}
\usepackage{fixltx2e}
\usepackage{stfloats}

\usepackage{amsmath,bm}
\usepackage{amssymb}
\usepackage{color}
\usepackage{subfig}

\usepackage{multirow}
\usepackage{float}
\usepackage{caption}
\usepackage[normalem]{ulem}
\usepackage{comment}
\usepackage{hyperref}
\usepackage{cite}
\usepackage[font=footnotesize]{subfig}

\usepackage{ifpdf}

\usepackage{epstopdf}
\usepackage[cmex10]{amsmath}

\usepackage{array}
\usepackage{mdwmath}
\usepackage{mdwtab}
\usepackage{eqparbox}
\usepackage{verbatim}
\usepackage{booktabs}
\usepackage{multirow}
\usepackage{graphicx}
\usepackage{algorithm}
\usepackage{algorithmic}
\usepackage{array} 
\usepackage{subcaption}  
\usepackage[colorlinks=true, allcolors=blue]{hyperref}  

\begin{document}
	\title{\LARGE \bf
Interactive OT Gym: A Reinforcement Learning-Based Interactive Optical tweezer (OT)-Driven Microrobotics Simulation Platform
}
\author{
Zongcai Tan, Dandan Zhang
\thanks{Zongcai Tan, Dandan Zhang are with the Department of Bioengineering, Imperial-X Initiative, Imperial College London, London, United Kingdom.  Corresponding: d.zhang17@imperial.ac.uk.}
}

\maketitle

\begin{abstract}
Optical tweezers (OT) offer unparalleled capabilities for micromanipulation with submicron precision in biomedical applications. However, controlling conventional multi-trap OT to achieve cooperative manipulation of multiple complex-shaped microrobots in dynamic environments poses a significant challenge. To address this, we introduce Interactive OT Gym, a reinforcement learning (RL)-based simulation platform designed for OT-driven microrobotics. Our platform supports complex physical field simulations and integrates haptic feedback interfaces, RL modules, and context-aware shared control strategies tailored for OT-driven microrobot in cooperative 
 biological object manipulation tasks. This integration allows for an adaptive blend of manual and autonomous control, enabling seamless transitions between human input and autonomous operation. We evaluated the effectiveness of our platform using a cell manipulation task. Experimental results show that our shared control system significantly improves micromanipulation performance, reducing task completion time by approximately 67\% compared to using pure human or RL control alone and achieving a 100\% success rate. With its high fidelity, interactivity, low cost, and high-speed simulation capabilities, Interactive OT Gym serves as a user-friendly training and testing environment for the development of advanced interactive OT-driven micromanipulation systems and control algorithms. 
 
 For more details on the project, please see our website https://sites.google.com/view/otgym
\end{abstract}

\section{Introduction}
Efficient, safe, and precise micromanipulation is crucial in biomedical applications such as cell sorting, tissue engineering, and microassembly~\cite{ref1}. Optical tweezers (OT), with their piconewton-level precision at micro-nanoscale~\cite{ref2}, are frequently used alongside microfluidic chips to manipulate and sort rare cells in small sample volumes~\cite{ref3,ref4,ref5}. However, for cells highly sensitive to phototoxicity and thermal damage, such as red blood cells, neurons, and circulating tumor cells (CTCs)\cite{ref6}, direct laser exposure can be unsafe during OT-based manipulation. Hence, manipulating microrobots via OT to achieve indirect cell handling is a more attractive solution \cite{butaite2019indirect,xie2018manipulation,hou2023review}.

Using multi-trap optical tweezers (OT) to control complex-shaped microrobots for indirect biological manipulation presents significant challenges, including precise alignment and stable cooperative control~\cite{chowdhury2013automated,ta2019stochastic,thakur2014indirect,ref7}. Precise force control is particularly crucial, as excessive force can damage delicate biological structures. Therefore, developing haptic feedback systems to enhance users' perception of force is essential~\cite{ref8,ref9,ref10,ref11,pacoret2013invited,1111}. Existing haptic devices primarily assist users in path following or obstacle avoidance, and the manipulated objects are typically simple geometric shapes such as spheres or ellipsoids~\cite{coffey2022collaborative,lee2021real}.

Furthermore, most robotic simulation platforms are designed for macro-scale systems, making them unsuitable for OT-driven microrobotics, which require adaptability and a force-sensitive human interaction interface. The lack of simulation platforms further limits accessibility and slows innovation. Additionally, autonomous control of OT-driven microrobots remains underexplored. While reinforcement learning (RL) has been used to improve efficiency and adaptability in optical manipulation~\cite{ref12,ref13}, it has not yet been applied to interactive control of complex-shaped microrobots or the unique challenges inherent in OT environments.

\begin{figure}[t]
  \captionsetup{font=footnotesize,labelsep=period}
\centering
\includegraphics[width=0.85\hsize]{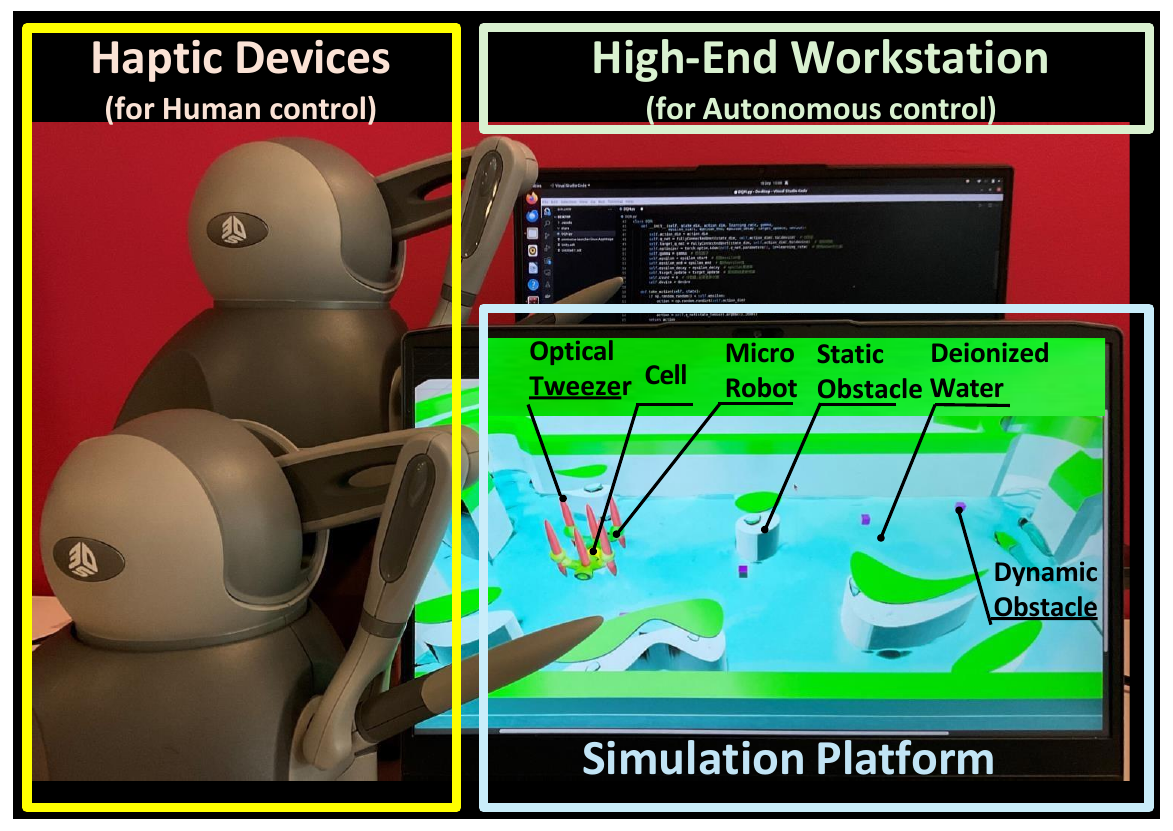}
\caption[The framework of shared control for OT micromanipulation]{Overview of the Interactive OT Gym Components: Haptic Devices for Human Control, High-End Workstation for Autonomous Control, and Simulation Platform.}
\label{OVERrallsystem}
     \vspace{-0.6cm}
\end{figure}%
To overcome these limitations,we developed a simulation platform, which integrates physical principles, capable of simulating complex phenomena like fluid resistance, dynamic obstacles, and random thermal motion in OT environments. This high-fidelity platform supports the evaluation of autonomous, semi-autonomous, and shared control algorithms. However, in complex tasks, the limited range of optical traps and the challenges in stable long-distance navigation lead to high computational costs and long training times for training RL model. To address these issues, we designed a progressive RL algorithm that enhances path planning efficiency with A* and adaptability in dynamic control with RL, providing a data-efficient solution for optical microrobot autonomy~\cite{ref14}.

While full automation can enhance efficiency, ensuring safety and dexterity in micromanipulation within dynamic and uncertain biological environments remains a critical challenge~\cite{ref15}. Manual control provides immediate feedback through human real-time decision-making; however, it often suffers from low efficiency due to the complexity of micromanipulation tasks and operator fatigue in dynamic environments~\cite{ref16}.
Mela Coffey et al.~\cite{coffey2022collaborative} proposed a collaborative control framework that integrates manual control with automatic algorithms, but its fixed control weight allocation and switching time limit its adaptability in dynamic environments. Similarly, Quang Minh Ta et al.~\cite{ta2021human} introduced a haptic feedback system, but operators must simultaneously manage obstacle avoidance and navigation speed, which increases their cognitive load and limits automation. 
In contrast, this paper presents a context-aware shared control strategy that combines RL-driven autonomous control with enhanced haptic feedback, dynamically adjusting control weights between human and robot in response to environmental changes. This approach offers a promising solution by improving safety, flexibility, and efficiency in micromanipulation systems.

To this end, this platform integrates a RL autonomous control system and a haptic feedback-enhanced teleoperation system, achieving for the first time the combination of manual and autonomous control in a simulation environment of OT and microfluidic chips to support human-robot collaboration, thereby improving the safety and efficiency of micromanipulation.

In summary, to address the current limitations in OT-driven microrobotics, we propose the \textbf{Interactive OT Gym}, the first simulation platform (to the best of our knowledge) that supports complex physical field simulations. This platform offers high-fidelity force feedback and an environment for training and deploying reinforcement learning (RL) models for autonomous microrobot control.
The \textbf{main contributions} of this paper are as follows:
\begin{itemize} 
\item \textbf{High-fidelity, cost-effective RL training environment}: The platform provides a realistic and interactive simulation environment that is both low-cost and capable of rapid testing. 
\item \textbf{Integration of shared control strategies}: Interactive OT Gym uniquely combines RL-based autonomous control with haptic feedback-enhanced teleoperation. 
\end{itemize}
Furthermore, unlike  previous platforms that only support autonomous algorithm simulation, our platform includes a haptic feedback teleoperation interface and can realistically reproduce complex physical phenomena in OT micromanipulation environments, such as fluid resistance, dynamic obstacles, and random thermal motion.

\begin{figure*}[htbp!]
  \captionsetup{font=footnotesize,labelsep=period}
\centering
\includegraphics[width=1\hsize]{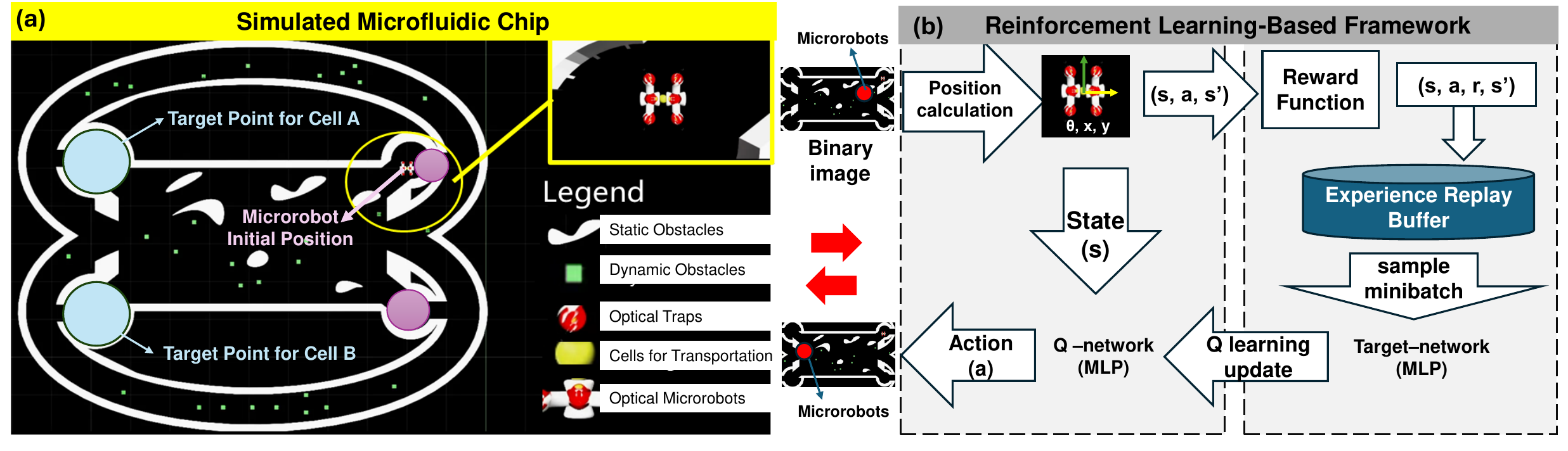}
\caption[The framework of shared control for OT micromanipulation]{The distributed architecture of shared control (connected by ROS), the motion of microrobots is determined by both RL control and human teleoperation. (a) Simulated Microfluidic Chip (b) Hardware \& Interactive environment \& DQN-Optimized Speed Control for Microrobots. }
\label{fig-sharedOverall}
    \vspace{-0.5cm}
\end{figure*}%

\section{Methodology}

\subsection{System Overview}


The Interactive OT Gym was designed as a distributed architecture-based simulation platform for OT-driven microrobot collaborative manipulation, which was constructed by a pair of Geomagic Touch haptic device, high-end workstation, and a simulation environment (see Fig. \ref{OVERrallsystem}). This integrated system is composed of three main modules:

\begin{itemize} \item \textbf{Simulator:} A high-fidelity physical simulation environment that provides accurate motion modeling and visual rendering.  (See Fig.\ref{fig-sharedOverall} (a))
\item \textbf{Autonomous Control:} A navigation module based on reinforcement learning (RL) for autonomous micromanipulation. (See Fig.\ref{fig-sharedOverall} (b))
\item \textbf{Manual Control:} An interactive teleoperation interface enhanced with haptic feedback for precise and intuitive user control. 
\end{itemize}

Each module is deployed on a separate device, with communication managed by the Robot Operating System (ROS). This distributed architecture prevents resource contention by enabling parallel execution and ensuring computational efficiency. In this study, we use a complex-shaped optical microrobot developed by Zhang et al.~\cite{ref7} as an example.

\subsection{Construction of the Physical Simulation Environment}

The high cost and complexity of traditional physical experiments limit the training and validation of algorithms. To accelerate this process and save time, we developed a virtual environment using NVIDIA Omniverse's Isaac Sim physics engine. This environment accurately simulates a micromanipulation system that combines OT and microfluidics, which can be used for RL model training. More specifically, we created a simulated microfluidic environment (Fig.\ref{fig-sharedOverall} (a)), in which optical traps are simulated to control microrobots for indirect cell manipulation. The simulated microfluidic chip features a four-branch structure with two inlets and two outlets, and microchannels measuring 25 µm in width and 50 µm in height, filled with buffer fluid to ensure stability. The key forces acting on the objects, as described by Equation~\eqref{eq:total_force}, include optical trapping $\mathbf{F}_{\text{optical}}$, Brownian motion $\mathbf{F}_{\text{Brownian}}$, van der Waals forces $\mathbf{F}_{\text{van der Waals}}$, viscous drag $\mathbf{F}_{\text{drag}}$, hydrodynamic forces $\mathbf{F}_{\text{hydrodynamic}}$, contact forces (including collision force)  $\mathbf{F}_{\text{contact}}$, and other forces $\mathbf{F}_{\text{other}}$. Additionally, four randomly moving obstacles are included to increase the training complexity.
\begin{equation}
\small
\begin{aligned}
\mathbf{F}(t) &= \mathbf{F}_{\text{optical}} + \mathbf{F}_{\text{hydrodynamic}} + \mathbf{F}_{\text{van der Waals}} + \mathbf{F}_{\text{drag}} \\
&\quad + \mathbf{F}_{\text{Brownian}} + \mathbf{F}_{\text{contact}} + \mathbf{F}_{\text{other}}
\end{aligned}
\label{eq:total_force}
\end{equation}

\subsection{Autonomous Control with Progressive RL Navigation}

\begin{figure}[htbp!]
  \captionsetup{font=footnotesize,labelsep=period}
\centering
\includegraphics[width=1.05\columnwidth]{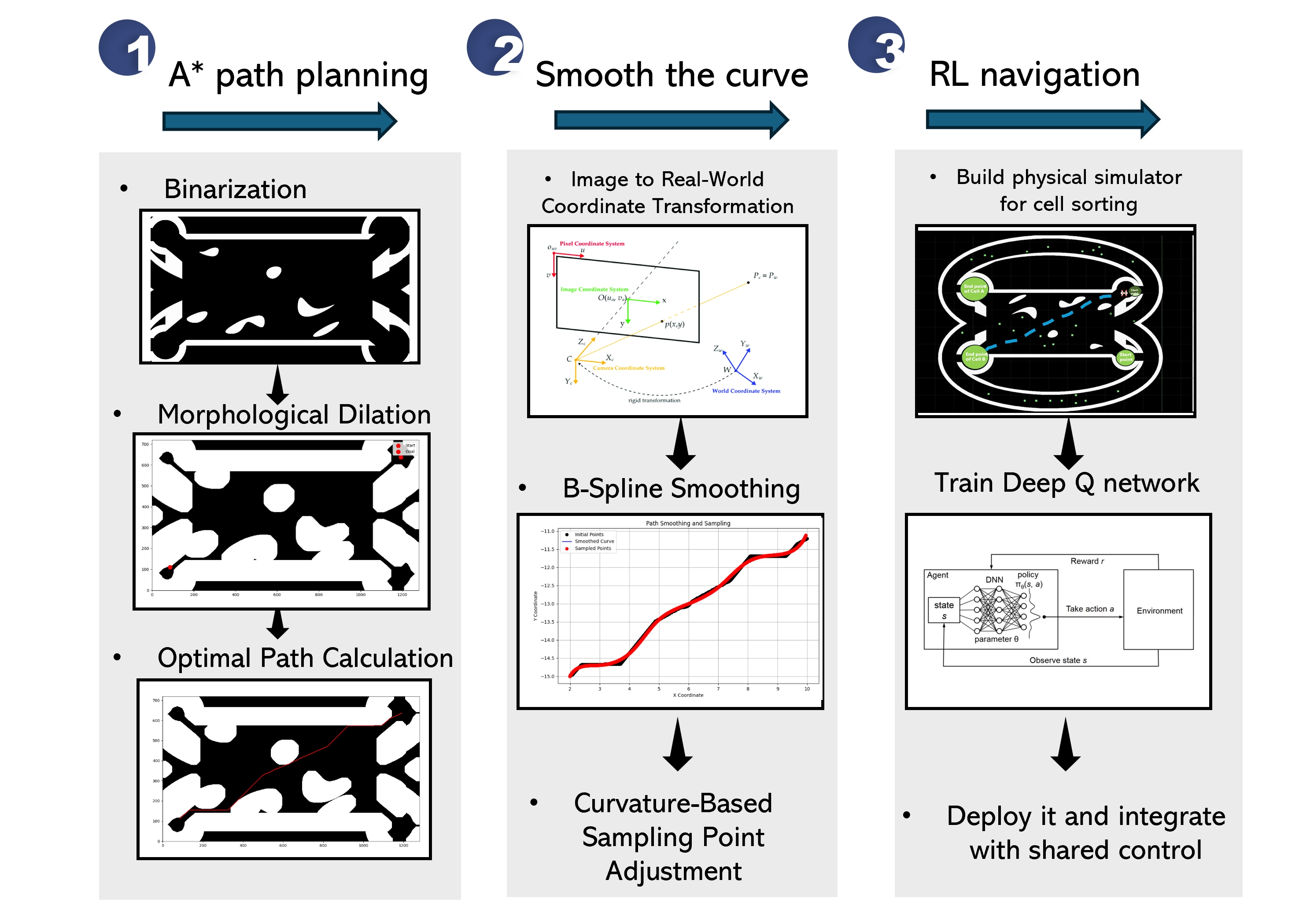}
\caption[Progressive training process for autonomous navigation]{Progressive training process for autonomous navigation.}
\label{fig-sharedGradual}
     \vspace{-0.4cm}
\end{figure}%

We proposed a progressive training strategy (Fig.~\ref{fig-sharedGradual}) that integrates A* path planning with RL-driven speed control to achieve fully autonomous navigation.
In this approach, A* is used for long-range path planning due to its computational efficiency and stability, as RL-based path planning would impose prohibitively high training costs. Meanwhile, precise speed control is essential in the microenvironment, where the limited range of optical trapping fields and environmental dynamics present significant challenges. Excessive speed may lead to cell or robot loss, whereas slower speeds exacerbate Brownian motion-induced disturbances and reduce navigation efficiency. Since directly modeling the relationship between navigation speed and dynamic environmental factors is impractical due to its complexity and uncertainty, RL provides a feasible solution for real-time speed adaptation, while A* ensures stable path planning. This hybrid approach reduces computational burden and enhances training efficiency in complex navigation scenarios.

In the preprocessing stage, grayscale images are binarized to separate obstacles from channels. To ensure safety, morphological dilation expands obstacle regions based on the microrobot’s size, taking into account both the robot's occupied volume and potential fine-tuning of the path, thus providing sufficient collision avoidance margins. The image coordinates are then converted to physical map coordinates for planning.

During path planning, the A* algorithm, extended with eight-directional movement, generates a more flexible pathfinding around obstacles, while B-spline interpolation smooths the path, eliminating sharp turns for smoother robot motion.
After path planning, an RL agent performs local speed control training along the A*-generated path. As shown in Fig.\ref{fig-sharedOverall} (b), the DQN framework uses a state space $s$ that includes the robot's real-time coordinates, cell positions, and obstacles. The action space $a$ consists of discrete speed levels, and the Q-network is updated using mini-batch sampling from the experience replay buffer $D$, following Equation~\eqref{eq:q_update}:
\begin{equation}
\small
Q(s_t, a_t) = Q(s_t, a_t) + \alpha \left[ r_t + \gamma \max_a Q(s_{t+1}, a) - Q(s_t, a_t) \right]
\label{eq:q_update}
\end{equation}

Here, $s_t$ is the state at time $t$, $a_t$ is the action, $r_t$ is the immediate reward, $\alpha$ is the learning rate, and $\gamma$ is the discount factor. The reward function $R(s, a)$ balances contact force optimization, collision avoidance, and speed control:
\begin{equation}
\small
R(s, a) = R_{\text{contact-force}}(s, a) + R_{\text{collision-penalty}}(s, a) + R_{\text{speed}}(s, a)
\label{eq:reward_function}
\end{equation}
where $R_{\text{contact-force}}$ rewards the agent for maintaining an optimal contact force, avoiding both excessive force, which could damage the cell, and insufficient force, which could lead to cell detachment, $R_{\text{collision-penalty}}$ penalizes collisions with obstacles or loss of the optical trap, and $R_{\text{speed}}$ encourages maintaining a high but safe speed along the path.

During training, an epsilon-greedy strategy is used to balance exploration and exploitation, with epsilon decaying from 1.0 to 0.01 over time. The target network parameters are updated periodically to ensure stable convergence.

Training was conducted on an NVIDIA GeForce RTX 4060 Laptop GPU (Nvidia Corporation, USA) \text color{red}{with 8GB of GDDR6 memory}, for 15 hours across 1000 episodes, with each episode consisting of approximately 200 time steps.

\subsection{Manual Control with Haptic Feedback}

Based on the physical simulation environment, this study developed a bimanual haptic feedback system for optical tweezer-driven microrobot teleoperation. The system enhances the user’s perception of micro-scale forces during indirect cell manipulation, preventing excessive pressure that could damage cells and ensuring the stability and safety of OT operations.

In this system, two haptic devices (Geomagic Touch, 3D Systems, USA) are used as input devices. The velocity of the device's tips incrementally controls the positions of two optical traps, which drive two optical microrobots. A vision-based force sensor measures the relative distance and direction between the robots and the tweezers' focal points. The system calculates the three-dimensional optical forces based on the optical trap force model (Equation~\eqref{eq:optical_force}) and scales these forces to provide real-time feedback through the haptic devices. 
\begin{equation}
\small
f_{\text{OT}}(X) = 
\begin{cases}
K \cdot \| X_{\text{trap}} - X_{\text{Object}} \|, & \text{if} \ \|X_{\text{trap}} - X_{\text{Object}}\| < \delta \\
C + \frac{A}{\| X_{\text{trap}} - X_{\text{Object}} \|^2}, & \text{if} \ \|X_{\text{trap}} - X_{\text{Object}}\| \geq \delta
\end{cases}
\label{eq:optical_force}
\end{equation}
where $X_{\text{trap}}$ and $X_{\text{Object}}$ denote the positions of the optical trap and the microrobot, respectively. $K=0.455$ is the trap stiffness, $\delta=0.446$ defines the threshold between near and far field interactions, $A=0.058$ scales the attractive force in the far field, and $C=0.01$ serves as an offset for distant interactions.
To enhance stability and improve user experience, the optical forces are empirically scaled. Both high-frequency noise from haptic feedback and disturbances from the operator’s manual input are filtered using low-pass filters. These noise sources, primarily caused by Brownian motion and human tremor, can lead to jittery control.


\subsection{Human-Robot Collaborative control}

\begin{figure}[htbp!]
  \captionsetup{font=footnotesize,labelsep=period}
\centering
\includegraphics[width=0.95\columnwidth]{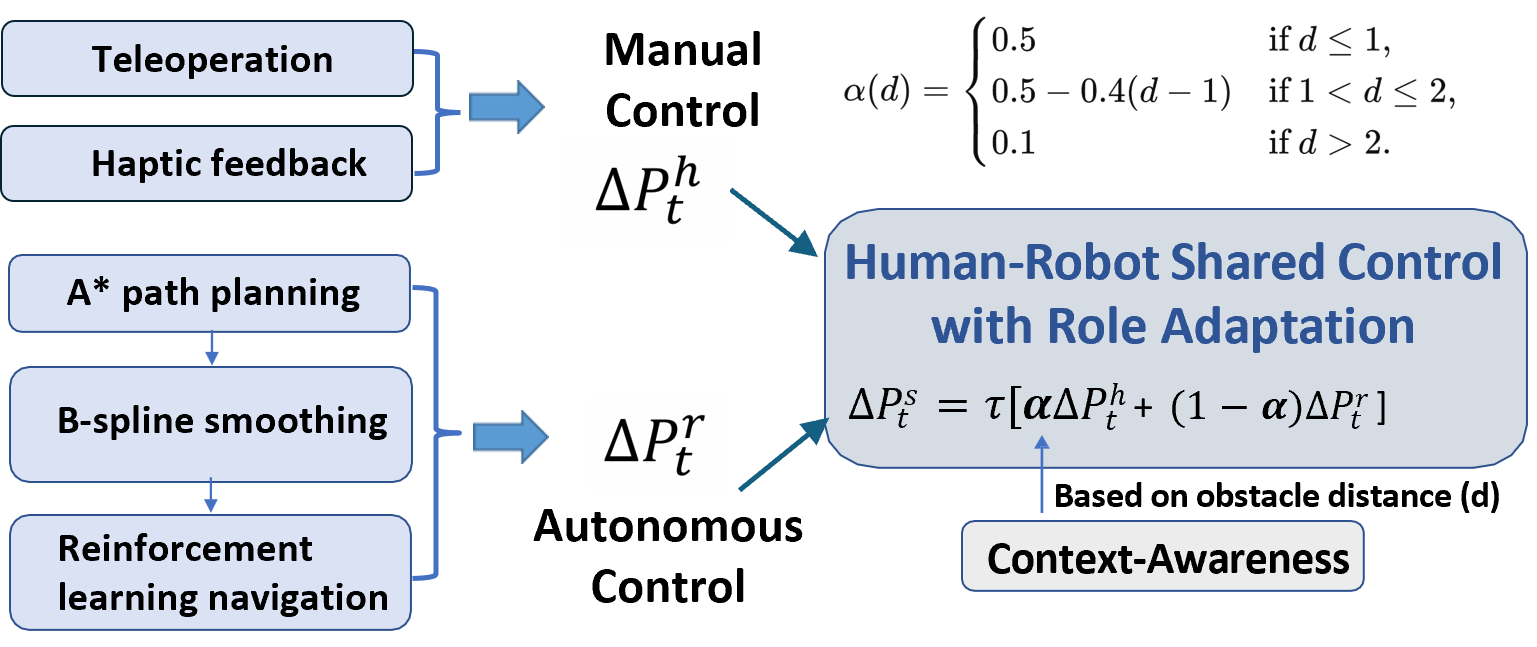}
\caption[Framework for context-aware adaptation]{Framework for human-robot shared control based on context-aware adaptation. Here we substitute $d_1=1, d_2=2$ into Equation~\eqref{eq:alpha_formula}.}
\label{fig-sharedcontrol111}
     \vspace{-0.2cm}
\end{figure}

To enable safer manipulation in complex microscopic environments, a shared control strategy was developed based on real-time distance adjustments (see Fig. \ref{fig-sharedcontrol111}). The system dynamically adjusts the weight distribution between human control and automated navigation by calculating the real-time distance between the robot and dynamic obstacles, offering three operation modes: fine-tuning, balanced, and autonomous.

Specifically, the displacement increment of the robot control is defined as $\Delta P_t^s$, where $\Delta P_t^r$ is determined by the automated navigation algorithm, and $\Delta P_t^h$ is the control command from the human operator. The motion scaling factor is $\tau$, and the weight parameter $\alpha$ is dynamically adjusted based on the real-time distance $d$. The control formula is $\Delta P_t^s = \tau [\alpha \Delta P_t^h + (1 - \alpha) \Delta P_t^r]$, where the value of $\alpha$ is dynamically adjusted according to the distance $d$, as defined by:

\begin{equation}
\small
\alpha = 
\begin{cases} 
    0.5 & \text{if } d \leq d_1, \\
    0.5 - 0.4 \times \frac{d - d_1}{d_2 - d_1} & \text{if } d_1 < d \leq d_2, \\
    0.1 & \text{if } d > d_2.
\end{cases}
\label{eq:alpha_formula}
\end{equation}

This shared control strategy ensures accurate fine manipulation while enhancing automation efficiency, providing an effective and safe solution for OT-driven microrobot-assisted cell sorting tasks in complex environments.

\section{Experiments and Results}

\subsection{Evaluation of Path Planning Performance}

\subsubsection{Task Description}

We evaluated the performance of manual and algorithm-generated trajectories using the motion capture module of our proposed Interactive OT Gym. OT controlled two microrobots to grasp and transport cells from a starting point to a destination in a cooperative manner. The platform recorded the trajectories, and we conducted qualitative and quantitative analyses—including frequency domain smoothness analysis—to assess manual paths, A* algorithm-generated paths, and B-spline smoothed paths based on A*.

\subsubsection{Results and Analysis}
\begin{figure*}[htbp]
  \captionsetup{font=footnotesize,labelsep=period}
\centering
\includegraphics[width=1\hsize]{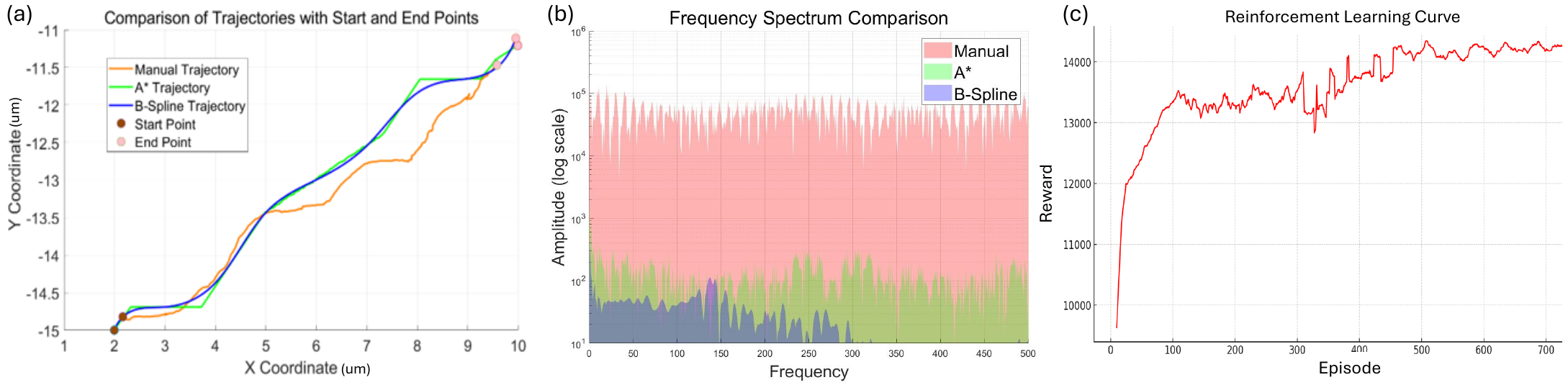}
\caption{ (a) Visualization of the three paths: manual, A*, and B-spline smoothed paths. (b) Spectral analysis results of the three paths: manual, A*, and B-spline smoothed paths. (c) learning curve of cumulative reward.}
\label{fig-pathQuantity}
    \vspace{-0.4cm}
\end{figure*}

Fig.~\ref{fig-pathQuantity} (a) visualizes the three paths. The manual path shows noticeable detours and jitter; the A* path has rigid turns; the B-spline smoothed path demonstrates a smoother trajectory closely approximating the shortest path.

\begin{table}[ht]
  \captionsetup{font=footnotesize,labelsep=period}
\centering
\caption{Comparison of different trajectory types using selected metrics.} 
\scalebox{0.9}{
\begin{tabular}{|l|r|r|r|}
\hline
Trajectory Type & Total Path Length & Mean Curvature & Angular Deviation \\
\hline
Manual    & 9.6193 & 314.34   & 45.962 \\
A*        & 9.4362 & 3.3893   & 9.5632 \\
B-Spline  & 9.1352 & 0.64264  & 0.30768 \\
\hline
\end{tabular}
}
\label{tab:trajectory_comparison}
\end{table}

Quantitative analysis (Table~\ref{tab:trajectory_comparison}) demonstrates that the B-spline path outperforms the others in total path length, mean curvature, and angular deviation. It achieves a shorter path and eliminates the sharp turns of the A* path through smoothing, resulting in a more continuous trajectory. The A* algorithm, constrained by discrete movements, introduces high curvature at turns, which is significantly reduced by the B-spline smoothing.

Frequency domain analysis (Fig.~\ref{fig-pathQuantity} (b)) shows:
\begin{itemize}
    \item the manual path contains significant high-frequency components, indicating substantial jitter and noise.
    \item the A* path reduces some high-frequency elements but still retains some noise.
    \item the B-spline smoothed path effectively eliminates high-frequency noise, confirming its superior smoothness.
\end{itemize}

The B-spline smoothed path performs best across all evaluation metrics. Its shorter path length, smooth curves, and low noise make it highly suitable for applications requiring high-precision path planning.

\subsection{Evaluation of RL Training Effectiveness}

\subsubsection{Experiment Description}

This experiment aimed to optimize microrobot speed control along a predefined trajectory using RL to balance the navigation navigation efficiency and safety in OT micromanipulation.
The RL model focused on optimizing speed control along a B-spline smoothed trajectory, minimizing inefficiency from slow speeds and fluctuations caused by Brownian motion, while preventing trap loss or cell detachment at high speeds. The reward function included penalties for contact forces above 10pN and deviations of more than 0.2µm from the trap center, as well as rewards for selecting higher speed levels.

As shown in Fig.~\ref{fig-pathQuantity} (c), the RL model's reward values steadily increased and converged after around 700 episodes, indicating that the model successfully learned to balance speed and navigation precision.

\begin{figure}[htbp!]
  \captionsetup{font=footnotesize,labelsep=period}
\centering
\includegraphics[width=1.0\columnwidth]{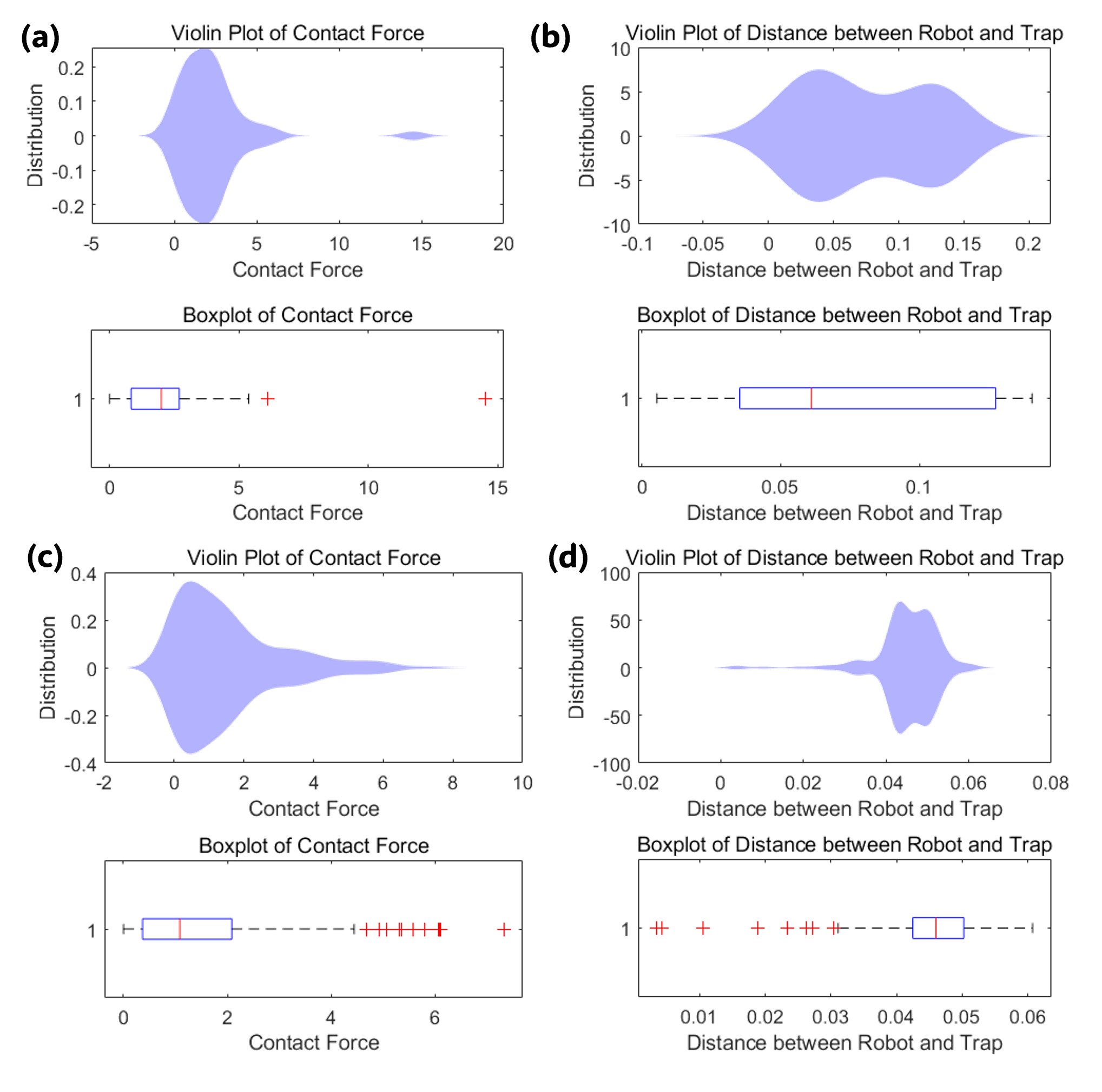}
\caption{Safety evolution during RL-based robot navigation training: distributions of contact force and robot-trap distance. (a), (b) before training; (c), (d) after training.}
\label{fig-rl1}
    \vspace{-0.4cm}
\end{figure}

Initially, violin and box plots (Fig.~\ref{fig-rl1} (a) and (b)) show frequent contact forces above 10pN, and median robot distance from the trap center exceeded 0.06µm, with some values approaching the 0.2µm threshold. However, as training progressed, the RL agent improved its strategy. By the end of training, contact forces remained below 10pN, and the robot stayed within 0.06µm of the trap center (Fig.~\ref{fig-rl1} (c) and (d)), demonstrating that RL training effectively reduced high-risk behaviors, improving stability and safety.

RL training successfully optimized speed control, reduced contact force fluctuations, and improved navigation precision. By balancing speed and safety, the model achieved stable and efficient control in complex micromanipulation tasks, demonstrating strong potential for applications requiring safety and rapid operations.

\subsection{Evaluation of RL Dynamic Speed Control Performance}

\subsubsection{Experiment Description}

We evaluated the performance of RL dynamic speed control by comparing it to a constant speed control using six speed levels, ranging from 0.27µm/s to 0.51µm/s, with speeds increasing from group 1 to group 6. The metrics for comparison included task completion time and success rate (the percentage of the total distance completed).

\subsubsection{Results and Analysis}

\begin{table}[htbp]
\captionsetup{font=footnotesize,labelsep=period}
\centering
\caption{Average Success Rates (ASR) and Task Completion Times (ATCT) for RL-based control and six groups of constant speed control}
\label{table:success_task_times}
\renewcommand{\arraystretch}{1.0}  
\setlength{\tabcolsep}{6pt}  
\scriptsize  
\begin{tabular}{|c|c|c|c|c|c|c|c|}
\hline
\textbf{Metric} & \textbf{RL} & \textbf{G1} & \textbf{G2} & \textbf{G3} & \textbf{G4} & \textbf{G5} & \textbf{G6} \\ \hline
\textbf{ASR} & 1.000 & 0.209 & 0.689 & 0.836 & 1.000 & 1.000 & 1.000 \\ \hline
\textbf{ATCT (s)} & 38.92 & / & 36.20 & 41.41 & 44.80 & 53.40 & 66.56 \\ \hline
\end{tabular}
\vspace{0.2em}  
\parbox{0.9\linewidth}{\scriptsize \textit{Note:} `G' indicates 'Group'. The slash (/) indicates that all experiments in the group failed, so no task completion time is available for statistics.}
\vspace{-0.1cm}
\end{table}

As shown in Table~\ref{table:success_task_times}, the success rate of constant speed control decreased as the speed increased, eventually dropping below 21\%. Although slower speeds achieved a 100\% success rate, they significantly increased task completion time, making this approach inefficient. In contrast, RL-controlled dynamic speed maintained a 100\% success rate while significantly reducing task completion time. This was achieved through dynamic speed adjustment, which prevents loss of control over the microrobots at higher speeds.

The evaluation showed that RL dynamic speed control outperformed constant speed control by maintaining a higher success rate and significantly reducing task completion time. This improvement stems from RL's ability to adaptively adjust navigation speed based on the dynamic environment. 

\begin{figure*}[htbp]
  \captionsetup{font=footnotesize,labelsep=period}
\centering
\includegraphics[width=1\hsize]{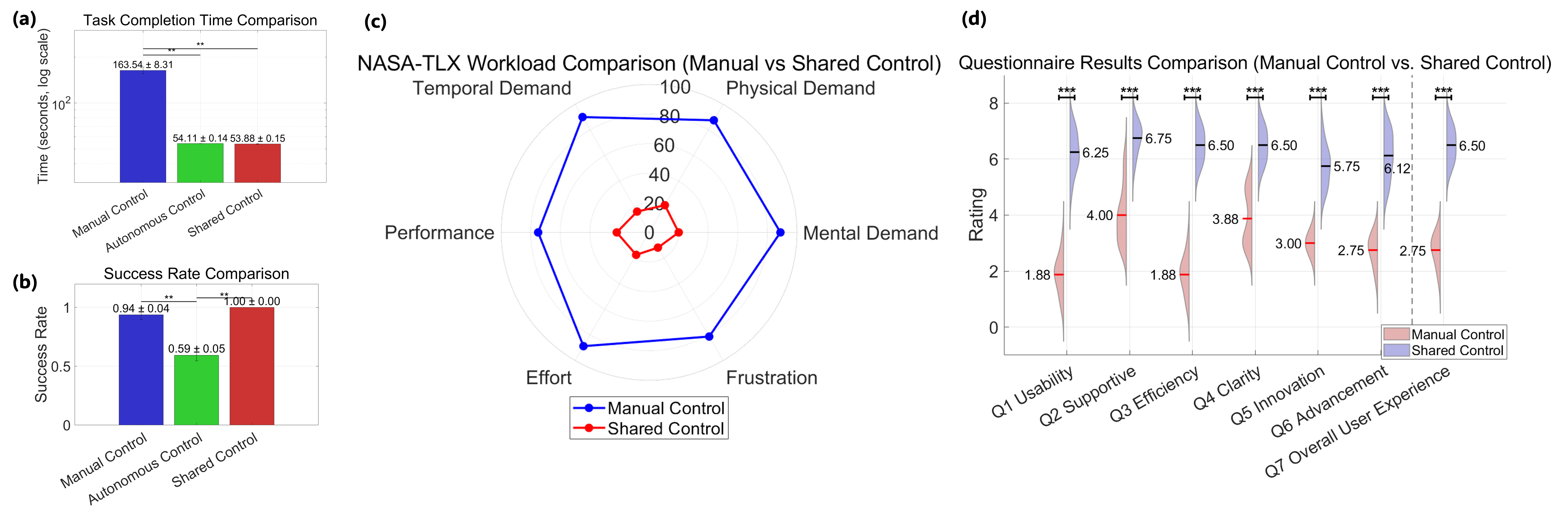}
\caption{ Comparison of success rate (a) and task completion time (b) under manual, autonomous, and shared control modes. (c) NASA-TLX scores under manual and shared control modes. (d) UEQ-S scores for manual and shared control modes. Statistical significance between groups is indicated by stars, where one star (\(*\)) represents a significant difference with \(p < 0.05\), two stars (\(**\)) represent \(p < 0.01\), and three stars (\(***\)) represent \(p < 0.001\).}
\label{fig-RLQuantity}
    \vspace{-0.5cm}
\end{figure*}

\subsection{Evaluation of Shared Control Effectiveness}

We conducted experiments to validate the performance of the OT shared micromanipulation system in terms of task efficiency and user experience, comparing it with manual and autonomous (RL-based) control.

\subsubsection{Experimental Setup}

Eight volunteers (ages 22–27, two females and six males) from Imperial College London participated. Five had prior experience with micromanipulation devices, and four had used haptic feedback devices. The task involved using an optical microrobot to grasp cells and navigate through a microfluidic chip with dynamic obstacles, transporting cells from start to destination.

\begin{figure}[htbp]
  \captionsetup{font=footnotesize,labelsep=period}
\centering
\includegraphics[width=1\columnwidth]{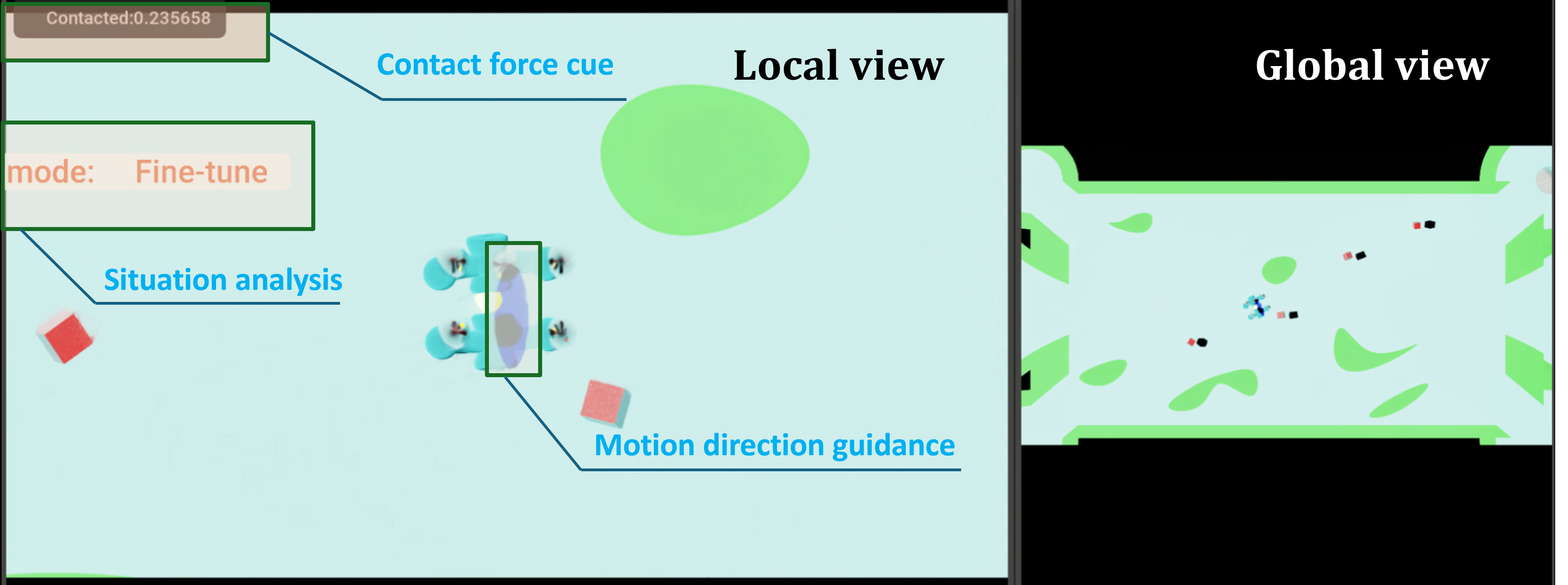}
\caption{User interface during microrobot manipulation, showing local and global views with situational awareness cues, haptic force, and motion direction guidance.}
\label{fig-Userinter}
    \vspace{-0.2cm}
\end{figure}

Participants initially operated the OT via a Geomagic Touch device, receiving haptic feedback for haptic perception of the gripping force. After grasping the cells, the RL algorithm handled navigation, with users making fine adjustments when needed. As the microrobot approached dynamic obstacles, the manual control weight gradually increased through a context-aware adaptation algorithm, prompting users through visual cues from the user interface for precise "fine-tune" interventions (Fig.~\ref{fig-Userinter}). Haptic feedback adjusted resistance to prevent loss of control or cell detachment if users moved too quickly.

\subsubsection{Quantitative Results}

Key metrics were task completion time and success rate (the proportion of dynamical obstacles avoided). Each mode (manual, autonomous, shared) was tested eight times for reliability.

According to Fig.~\ref{fig-RLQuantity} (a), manual control had the longest completion time ($163.5 \pm 8.3$~s), while autonomous and shared control times were much shorter ($54.1 \pm 0.1$~s and $53.9 \pm 0.2$~s, respectively). ANOVA confirmed manual control was significantly slower than the other modes ($p < 0.01$).
According to Fig.~\ref{fig-RLQuantity} (b), shared control achieved a 100\% success rate, outperforming manual ($94\% \pm 4\%$) and autonomous control ($59\% \pm 5\%$), which struggled with dynamic obstacles. Shared control combined RL-based global navigation with user intervention, improving overall performance.

\subsubsection{User Experience Evaluation}

After the experiments, participants completed the NASA Task Load Index (NASA-TLX) and User Experience Questionnaire Short version (UEQ-S) to assess workload and user satisfaction. As shown in Fig.~\ref{fig-RLQuantity} (c), shared control significantly reduced workload across all six NASA-TLX dimensions compared to manual control, especially in mental and physical demand, highlighting the reduced user burden due to task delegation to automation. The UEQ-S results (Fig.~\ref{fig-RLQuantity} (d)) showed shared control scored higher across all seven dimensions, particularly in supportiveness and effectiveness ($p < 0.001$). Although manual control also scored well in supportiveness and clarity, shared control significantly improved overall user satisfaction.

In conclusion, the shared control system outperformed manual and autonomous modes by combining RL-based automation with human adaptability. It reduced task completion time, improved success rates, and lowered workload, demonstrating its potential for complex micromanipulation tasks requiring precision and safety.

\section{Conclusion}

In this paper, We introduced Interactive OT Gym, the first interactive simulation platform for OT-driven microrobotics that incorporates complex physical field simulations. By integrating a high-fidelity simulator, haptic feedback-based manual control modules, progressive RL-based autonomous control modules, and shared control strategies, we constructed and effective human-robot collaboration through seamless integration of manual and autonomous control.
Experimental results demonstrated that our shared control system significantly enhanced micromanipulation performance, reducing task completion time by 67\% and increasing the success rate to 100\%. User workload was also substantially reduced, highlighting the platform's effectiveness in improving efficiency, precision, and safety in OT-driven micromanipulation tasks in simulated cell transportation.
This simulation platform offers an efficient and low-cost training environment for the development of next-generation collaborative OT-driven microrobotics systems. It holds significant potential in areas such as tissue engineering, micro-assembly, and biological object manipulation, where precise and efficient micromanipulation is crucial.

Future work will focus on improving simulation fidelity to bridge the sim-to-real gap, enabling more accurate RL-based control models to be transferred to real-world biomedical applications after training in the simulated environment. 

\section*{Acknowledge}
This research was conducted in accordance with ethical guidelines and was approved by the Imperial College London Research Ethics Committee (Approval ID: 7134867).

\bibliographystyle{IEEEtran}

\bibliography{references}
\end{document}